%% file: main.tex
\documentclass[runningheads]{llncs}

\usepackage{eccv}

\usepackage{eccvabbrv}

\usepackage{graphicx}
\usepackage{booktabs}

\usepackage[accsupp]{axessibility}  

\usepackage{hyperref}

\usepackage{orcidlink}

\input{preamble.tex}

\begin{document}

\title{From Uncertainty to Stability and Fidelity: Guiding Sparse-View 3D Gaussian Splatting with Fisher Information}

\titlerunning{From Uncertainty to Stability: Guiding Sparse-View 3D Gaussian Splatting with Fisher Information}

\author{
Junbao Zhou\inst{1} \and
Qingshan Xu\thanks{Qingshan Xu is the corresponding author.}\inst{2} \and
Yuan Zhou\inst{1} \and
Xiaolong Shen\inst{3} \and
Beier Zhu\inst{2} \and
Kesen Zhao\inst{1} \and
Yiming Zeng\inst{4} \and
Chen Bai\inst{4} \and
Cheng Lu\inst{4} \and
Hanwang Zhang\inst{1}
}

\authorrunning{Junbao Zhou et al.}

\institute{
    Nanyang Technological University, Singapore \and
    University of Science and Technology of China, China \and
    Zhejiang University, China \and
    Xpeng Motors, China
}

\maketitle

\begin{figure*}[t]
    \centering
    \includegraphics[width=1.0\textwidth]{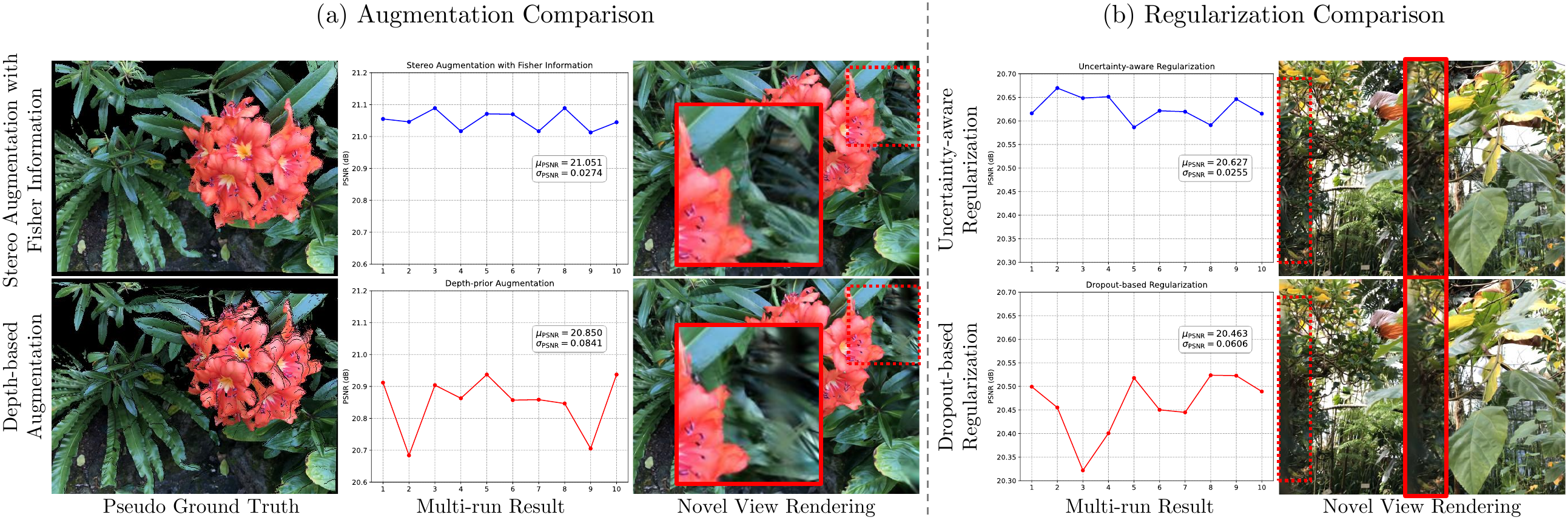}
    \caption{
        Comparison between our proposed strategies and traditional depth-based augmentation and Dropout-based regularization for sparse-view 3DGS. $\sigma_{\text{PSNR}}$ and $\mu_{\text{PSNR}}$ denote the standard deviation and the mean PSNR across multiple runs, respectively. \textbf{Fig. (a)} illustrates that curating pseudo ground truths by warping a single training view fails to fully utilize geometric priors, resulting noticeable hollow areas in pseudo ground truth. Randomly sampling pseudo ground truth viewpoints further introduces instability in optimization, leading to suboptimal performance. Our \textbf{Stereo Augmentation with Fisher Information} selects most informative views to curate high-quality pseudo ground truths, thus stabilizing the optimization ($\sigma_{\text{PSNR}} \downarrow : 0.084 \rightarrow 0.027$) and improving the rendering quality ($\mu_{\text{PSNR}} \uparrow : 20.87 \rightarrow 21.05$). \textbf{Fig. (b)} demonstrates the inherent randomness and instability of traditional Dropout-based regularization, which applys uniform dropout probability to all Gaussians. Our \textbf{Uncertainty-aware Regularization} adaptively adjusts the dropout probability of each Gaussian according to its uncertainty status, leading to more stable optimization ($\sigma_{\text{PSNR}} \downarrow : 0.061 \rightarrow 0.026$) and improved performance ($\mu_{\text{PSNR}} \uparrow : 20.46 \rightarrow 20.63$).
        }
    \label{fig:teaser}
\end{figure*}

\vspace{-0.6cm}

\input{0_abstract.tex}
\input{1_introduction.tex}
\input{2_related_works.tex}
\input{3_method.tex}

\input{4_experiment.tex}
\input{5_conclusion.tex}

\begin{figure*}
    \centering
    \includegraphics[width=0.99\textwidth]{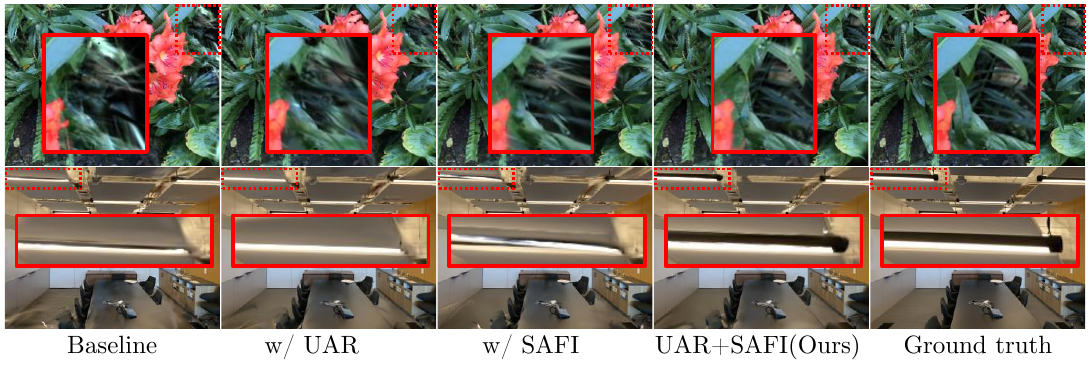}
    \caption{Qualitative ablation study on LLFF dataset in 3-view setting.}
    \label{fig:ablation}
\end{figure*}
\begin{figure*}
    \centering
    \includegraphics[width=0.99\textwidth]{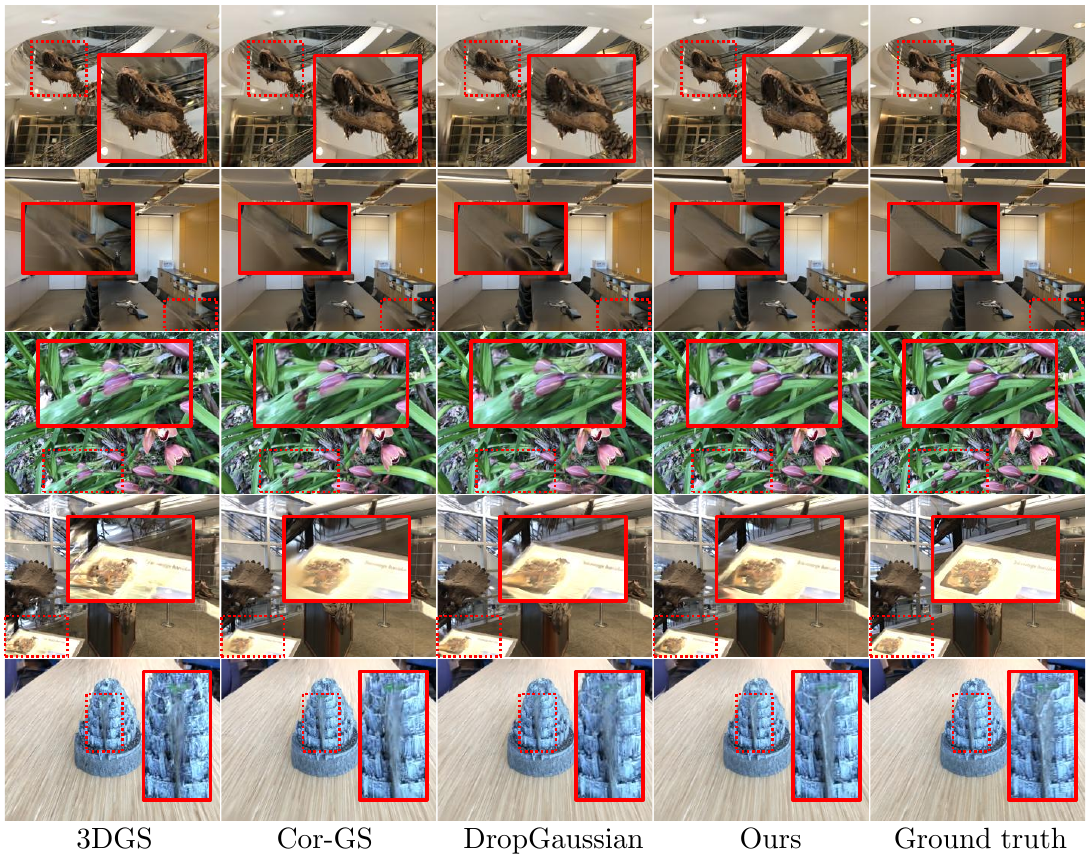}
    \caption{Qualitative comparisons with baseline methods on LLFF dataset (3-view).}
    \label{fig:llff_compare}
\end{figure*}
\begin{figure*}
    \centering
    \includegraphics[width=0.99\textwidth]{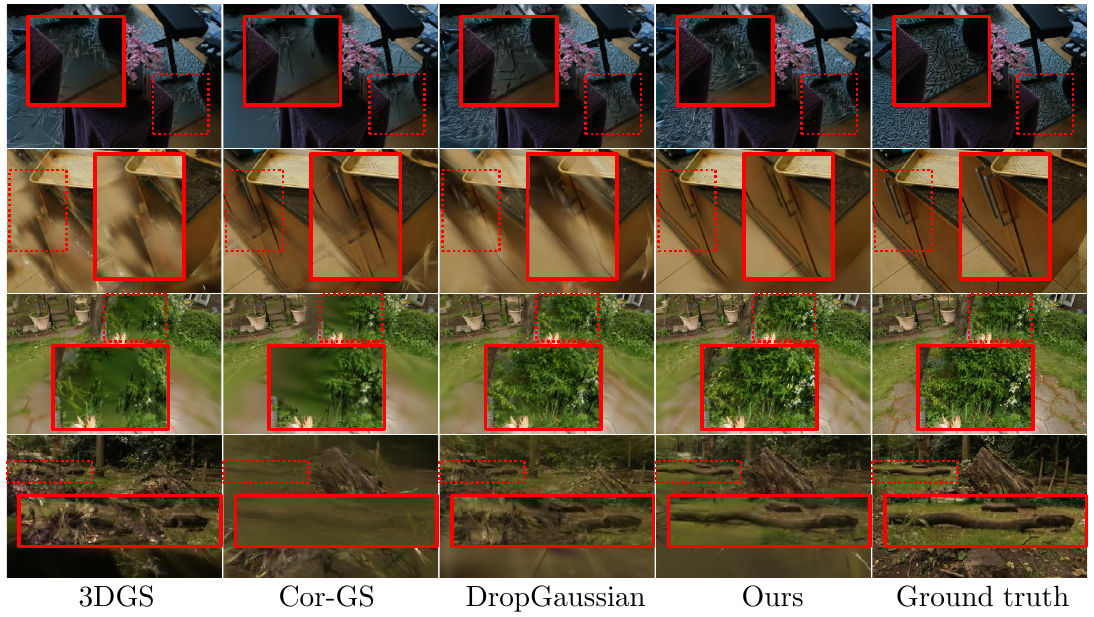}
    \caption{Qualitative comparisons with baseline methods on Mip-NeRF 360 dataset (12-view).}
    \label{fig:mip_compare}
\end{figure*}

%
%
\bibliographystyle{splncs04}
\bibliography{bibliography}



\end{document}

%% file: preamble.tex
\usepackage{tabularx}
\usepackage{booktabs}
\usepackage{pifont}
\usepackage{multicol}
\usepackage{multirow}
\usepackage{amsmath,amsfonts}
\usepackage{graphicx}
\usepackage{enumitem}
\usepackage{bm}
\usepackage{wrapfig}
\usepackage[table]{xcolor}

\newcommand{\cmark}{\ding{51}}

\newcommand\mypar[1]{\par\addvspace{1.0mm}\noindent\textbf{#1}\;}

\definecolor{tabfirst}{rgb}{1, 0.7, 0.7} 
\definecolor{tabsecond}{rgb}{1, 0.85, 0.7} 
\definecolor{tabthird}{rgb}{1, 1, 0.7} 

\newcommand{\bfx}{{\bm{x}}}

\newcommand{\bfy}{{\bm{y}}}
\newcommand{\bfY}{{\bm{Y}}}
\newcommand{\bftheta}{{\boldsymbol{\theta}}}

\newcommand{\Fisher}{{\mathcal{I}}}
\newcommand{\Expect}{{\mathbb{E}}}
\newcommand{\RealSet}{\mathbb{R}}

\newcommand{\Normal}{\mathcal{N}}
\newcommand{\entropy}{\mathbf{H}}
\newcommand{\Hessian}{\entropy''}
\newcommand{\IG}{\mathbf{IG}}
\newcommand{\trace}{\text{tr}}
\newcommand{\argmin}{\operatorname*{arg\,min}}

\newcommand{\data}{\mathcal{D}}

%% file: 0_abstract.tex
\begin{abstract}

3D Gaussian Splatting (3DGS) has emerged as a promising technique for novel view synthesis. However, 3DGS requires dense input views to achieve high-quality rendering. In sparse-view scenarios, 3DGS often prones to overfitting, resulting in noticeable artifacts and degraded rendering quality. Previous methods explore to address this issue by introducing additional priors (e.g. depth priors) or integrating regularization techniques (e.g. Dropout). 
However, these methods are often applied without principled guidance. In particular, prior-based augmentation typically samples novel viewpoints randomly, while Dropout-based regularization randomly removes Gaussians. The compounded randomness introduces uncertainty and instability, limiting the fidelity of novel view synthesis.
In this paper, we propose a novel method for sparse-view 3DGS that incorporates Fisher Information to quantitatively guide the utilization of geometric priors and regularization.
Specifically, our method comprises two key components: \textbf{(1) Stereo augmentation with Fisher Information.} By leveraging Fisher Information, we actively select most informative supporting views and use depth priors to curate reliable pseudo ground truths, which reduces randomness in augmentation and improves \emph{stability} and rendering \emph{fidelity}; \textbf{(2) Uncertainty-aware regularization.} We reduce the instability of Dropout-based regularization by using Fisher Information to quantitatively measure the uncertainty of each 3D Gaussian, and adaptively adjust the removal probability, leading to more \emph{stable} and effective regularization. With these two components, our method effectively mitigates overfitting and improves the \emph{stability} of optimization in sparse-view 3DGS, resulting in superior rendering \emph{fidelity}. Extensive experiments show that our method achieves state-of-the-art performance in sparse-view novel view synthesis benchmarks.


\end{abstract}

%% file: 1_introduction.tex
\section{Introduction}

Novel view synthesis (NVS) is an essential technique in computer vision and graphics, rendering realistic images from unseen viewpoints given a dense set of input images. Traditional methods use explicit 3D representations, such as point clouds \cite{Neural_rerendering} or meshes, to render novel views on 3D scenes. Recent years, learning-based methods, particularly Neural Radiance Fields (NeRF) \cite{nerf} and 3D Gaussian Splatting (3DGS) \cite{3DGS}, have drawn significant attention due to their ability to render photorealistic novel views. NeRF represents a scene using a continuous volumetric function parameterized by neural networks. More recently, 3DGS has emerged as an advanced approach that represents a scene with a set of 3D Gaussians, parameterized by their positions, colors, and opacities. Compared to NeRF, 3DGS can be rendered directly via rasterization, which leads to faster rendering speeds while maintaining high rendering quality of novel views.

\begin{wrapfigure}[12]{r}{0.45\textwidth}
    \vspace{-1.5\baselineskip} 
    \centering
    \includegraphics[width=\linewidth]{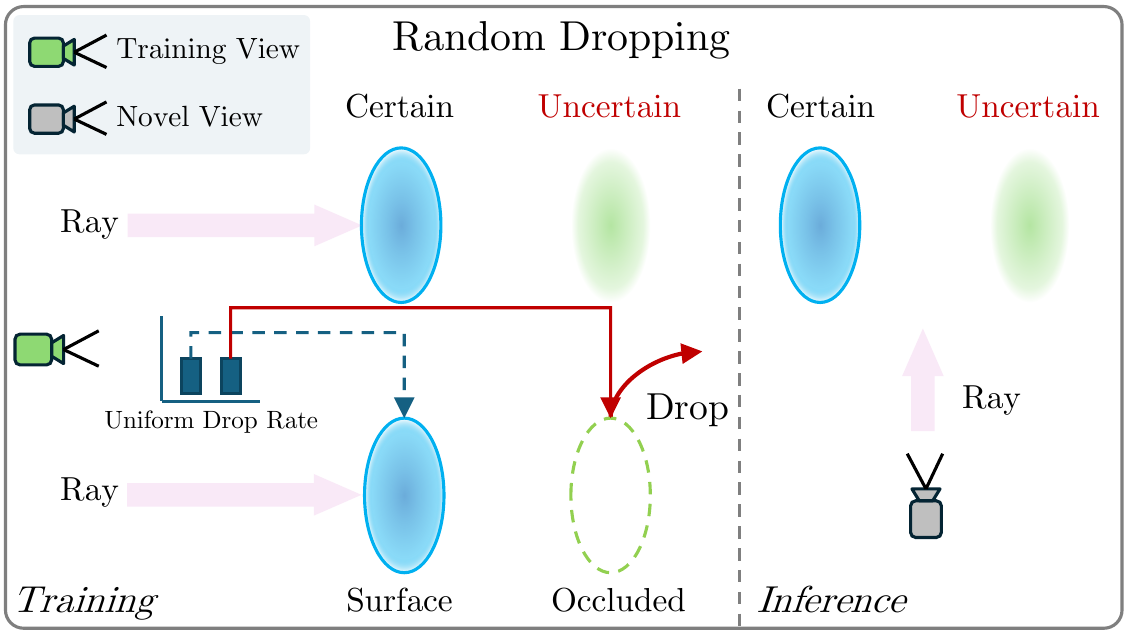}
    \vspace{-0.6cm}
    \caption{Uniform dropping rate possibly remove under-optimized Gaussians, while preserving over-optimized ones.}
    \label{fig:random_drop}
\end{wrapfigure}

However, 3DGS requires a large number of input views during training to achieve satisfactory rendering quality.
When restricted to a limited number of input views (i.e. sparse-view setting) \cite{sparsegs}, 3DGS suffers from drastic performance degradation due to insufficient geometric and appearance information. The lack of training information results in severe overfitting, harming the generalization ability of the rendering model in NVS.

To address the challenges in sparse-view scenarios, previous works have made efforts in two main directions:
\begin{enumerate}[leftmargin=*]
    \item \textbf{Prior-based augmentation.} \cite{dngaussian, fsgs, sparsegs} With the help of prior knowledge supplied by pre-trained models, training sets are augmented with curated pseudo ground truths. A common approach is to incorporate depth priors \cite{dngaussian, fsgs, sparsegs} produced by pre-trained depth estimation models.
    \item \textbf{Regularization techniques.} These methods regularize the training process with various constraints to prevent overfitting. For 3DGS, recent works \cite{dropgaussian, dropoutgs} have proven that Dropout is effective in sparse-view setting.
\end{enumerate}

However, these methods raise issues that limit the performance:
\begin{enumerate}[leftmargin=*]
    \item \textbf{Hollow and Random Augmentation.} Previous depth-based augmentation warps a single training view to curate pseudo ground truths, which fails to fully exploit stereo geometric priors, resulting in hollow areas in the augmented samples (Fig~\ref{fig:teaser}(a), bottom). Moreover, randomly sampling novel viewpoints introduces instability (Fig~\ref{fig:teaser}(a), bottom).
    \item \textbf{Regularization without Selectivity.} Dropout-based regularization applies a uniform removal probability without considering the status of each Gaussian. Consequently, surface Gaussians continue to receive larger gradient updates and become over-optimized, while occluded Gaussians remain under-optimized. When synthesizing novel views, the under-optimized Gaussians degrade the visual fidelity (Fig~\ref{fig:random_drop} and Fig~\ref{fig:teaser}(b), bottom).
\end{enumerate}




To address the issues above, we propose a novel method with Fisher Information guidance to fully utilize geometric priors and stabilize the optimization. Our method comprises two key components:

\begin{enumerate}[leftmargin=*]
    \item \textbf{Stereo Augmentation with Fisher Information.} To maximize the utility of depth priors, we leverage Fisher Information to actively select a set of most informative ground truth views. By warping these selected views, we curate high-quality pseudo-ground truths. This strategy augments the training set, supervising the model with robust geometric and appearance information.
    \item \textbf{Uncertainty-aware Regularization.} We primarily prioritize the removal of over-optimized Gaussians while protecting under-optimized ones. We utilize Fisher Information to quantitatively measure the uncertainty of each 3D Gaussian \cite{pup3dgs}, and adjust its dropout probability accordingly. Furthermore, we introduce \textbf{soft-scale dropout} to avoid the complete discarding of selected Gaussians, which retains the influence of dropped Gaussians and alleviates the \textit{over-compensation} of occluded Gaussians.
\end{enumerate}

As shown in Fig.~\ref{fig:teaser}, our proposed components effectively mitigates overfitting, improving the stability of optimization and the fidelity of novel view rendering in sparse-view 3DGS. Extensive experiments on three long-established benchmarks (LLFF, Mip-NeRF 360, and DTU) prove the superiority of our method, marking a new state-of-the-art in sparse-view NVS.

In summary, our contributions are as follows:

\begin{itemize}[leftmargin=*]
    \item We propose \textbf{Stereo Augmentation strategy with Fisher Information guidance} for sparse-view 3DGS that selects most informative views to curate high-quality pseudo ground truths with accurate geometric priors and augment the training data.
    \item We introduce \textbf{Uncertainty-aware Regularization} that measures the uncertainty of each 3D Gaussian and adaptively adjusts the dropout probability, improving stability and rendering quality.
    \item We further integrate \textbf{soft-scale dropout} into the \textbf{Uncertainty-aware Regularization} to avoid \textit{over-compensation} in Dropout.
\end{itemize}

%% file: 2_related_works.tex
\section{Related Works}

\subsection{Novel View Synthesis}

Novel View Synthesis (NVS) aims to generate unobserved views of a scene given a set of observed images. Traditional NVS methods employ explicit 3D representations, such as point clouds \cite{Neural_rerendering} or meshes \cite{A_comparison_and, modelingandrendering2023}.
Recent advances \cite{nerf,3DGS} in differentiable rendering have pushed the boundaries of NVS by achieving photorealistic quality.

\mypar{Neural Radiance Fields. }
NeRF \cite{nerf} represents a scene as a continuous volumetric function parameterized by a neural network. The neural network takes 3D coordinates $(x, y, z)$ and viewing direction $(\theta, \phi)$ as input and predicts color and density. Various methods have been proposed to improve rendering quality following the emergence of NeRF \cite{mip_nerf, mip_nerf_360, zip_nerf, aug_nerf, nerfren, light_field, ref_nerf, F2_nerf}. There are also works that focus on improving NeRF's efficiency by accelerating training and rendering \cite{tensorf, plenoxels, tri_miprf, neural_sparse_voxel, instant_neural_graphics, direct_voxel_grid, plenoctrees, point_nerf, fastnerf, kilonerf}.

\mypar{3D Gaussian Splatting. }
3DGS \cite{3DGS} advances differentiable rendering by achieving real-time speeds without losing quality in novel view synthesis. It represents a scene with a set of 3D Gaussians, parameterized by their positions, colors, and opacities. Unlike NeRF, which uses volumetric rendering coupled with a neural network, 3D Gaussians are rendered via rasterization pipeline, achieving  real-time speed. Subsequent works have further improved the rendering quality \cite{ges, 2d_gaussian_splatting, mini_splatting, absgs, pixel_gs} or reduced the memory consumption \cite{compact_3d_gaussian, compressed_3d_gaussian_splatting, compgs, hac, lightgaussian, reducing_the_memory_footprint}.

\subsection{Novel View Synthesis from Sparse Views}


Despite the impressive rendering quality, NeRF and 3DGS require a large number of input views during training. Under sparse-view conditions, these methods prone to severe overfitting. To address this problem, two main strategies have been explored: \textbf{Prior-based augmentation} and \textbf{Regularization.} 

\mypar{Prior-based Augmentation.}
Numerous works \cite{sparsenerf, depth_regularized, dngaussian, fsgs, sparsegs, depth_supervised_nerf, dense_depth_priors, darf, geoaug, diffusionerf, reconfusion, putting_nerf, clip} have introduced different types of priors to mitigate overfitting. The most common type is depth priors estimated by pre-trained depth models \cite{sparsenerf, depth_regularized, dngaussian, fsgs, sparsegs, depth_supervised_nerf, dense_depth_priors, darf, geoaug}.
GeoAug \cite{geoaug} warps a novel view to a training view with depth priors to enforce additional supervision. SparseSurf \cite{sparsesurf} augment the training set by curating pseudo ground truths from a single training view with depth priors.

\mypar{Regularization Techniques.}
Various NeRF-based methods have explored different regularization techniques, including depth regularization \cite{regnerf}, frequency regularization \cite{freenerf}, and geometric regularization \cite{sparf, geconerf}. As for 3DGS, recent works \cite{dropgaussian, dropoutgs} have proved that Dropout \cite{dropout} is an effective strategy to alleviate overfitting in sparse-view NVS.

However, these two main approaches either fail to fully utilize geometric priors or introduce instability, limiting the performance of sparse-view NVS. 

%% file: 3_method.tex
\section{Preliminaries}

\subsection{3D Gaussian Splatting}

3D Gaussian Splatting (3DGS) \cite{3DGS} represents a 3D scene with a set of 3D Gaussians $\{G_i\}$. Each Gaussian is defined as:

\begin{equation}
    G_i(x) = \exp\left(-\frac{1}{2}(x - \mu_i)^\top \Sigma_i^{-1} (x - \mu_i)\right)
\end{equation}
where $x \in \RealSet^3$ is a point in 3D space, $\mu \in \RealSet^3$ is the center of the 3D Gaussian, $\Sigma_i \in \RealSet^{3 \times 3}$ is the matrix defining the shape and orientation of the Gaussian, which can be decomposed as $\Sigma_i = R_i S_i S_i^\top R_i^\top$. $R_i \in \RealSet^{3 \times 3}$ is a rotation matrix, and $S_i \in \RealSet^{3 \times 1}$ is a scaling vector. In addition, each Gaussian also has an opacity parameter $\alpha_i \in [0, 1]$ and a color feature parameterized by spherical harmonic coefficients $SH_i \in \RealSet^{3 \times K}$. The actual color $c_i$ of Gaussian is computed based on the viewing direction and the spherical harmonic coefficients $SH_i$.

During rendering, the 3D Gaussians are firstly sorted based on their distance to the camera viewpoint. Then, the color of each pixel $p$ in 2D image is computed by blending all the Gaussians along the corresponding camera ray:
\begin{equation}
    C(p) = \sum_{i}^N c_i a_i(p) \prod_{j < i} (1 - a_j(p))
\end{equation}
where $N$ is the number of Gaussians contributing to pixel $p$, and $a_i(p)$ is the final opacity of Gaussian $G_i$ contributing to pixel $p$. $a_i(p)$ can be computed by:
\begin{equation}
    a_i(p)  = \alpha_i G_i^{2D}(p)
\end{equation}
where $G_i^{2D}(p)$ is obtained by projecting the 3D Gaussian $G_i(\bfx)$ onto the 2D image plane at pixel $p$.

\subsection{Fisher Information}

Given a probabilistic model $\bfY \sim p(\cdot \mid \bfx; \bftheta)$,
Fisher Information is useful in measuring how much does the distribution $p(\bfy \mid \bfx, \bftheta)$ shifts under a small change in $\bftheta$, i.e. the amount of information that an observation $(\bfx, \bfy)$ carries about the parameter $\bftheta$. Fisher Information is defined as:
\begin{equation}
\Fisher(\bftheta)
= \Expect_{\bfY \sim p(\cdot \mid \bfx, \bftheta)} \left[ \nabla_{\theta} \log p(\bfY \mid \bfx; \bftheta) \nabla_{\theta} \log p(\bfY \mid \bfx; \bftheta)^\top \right]
\end{equation}

\begin{figure}[ht]
    \centering
    \includegraphics[width=1\linewidth]{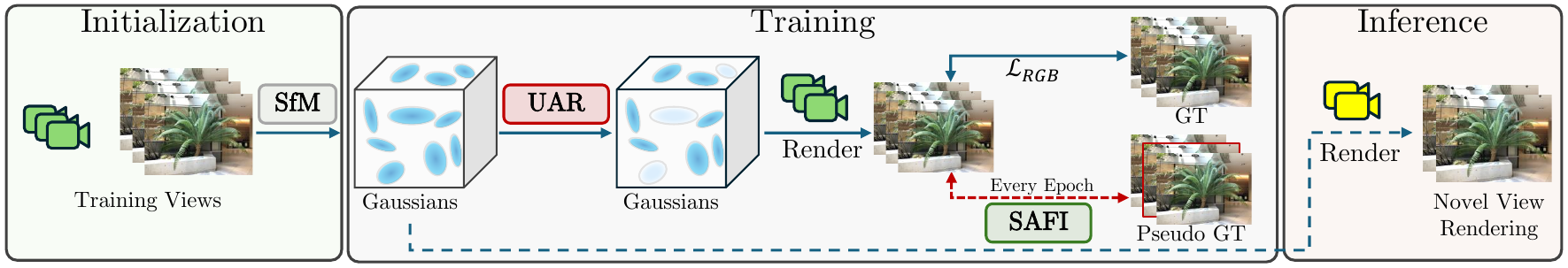}
    \vspace{-0.3cm}
    \caption{\textbf{Overall pipeline of our proposed method. } UAR denotes Uncertainty-aware Regularization, and SAFI denotes Stereo Augmentation with Fisher Information, which are two key components of our method. }
    \label{fig:pipeline}
    \vspace{-0.3cm}
\end{figure}

Under standard regularity conditions (Sec.~\ref{sec:fisher_regularity}), it is equivalent to:
\begin{equation}
    \begin{aligned}
        \Fisher(\bftheta)
         & = - \Expect_{\bfY \sim p(\cdot \mid \bfx; \bftheta)}
        \left[
            \nabla_{\theta}^2 \log p(\bfY \mid \bfx; \bftheta)
        \right]
        = \Expect_{\bfY \sim p(\cdot \mid \bfx; \bftheta)}
        \left[
            \Hessian[\bfY \mid \bfx; \bftheta]
        \right]
    \end{aligned}
    \label{eq:fisher_hessian}
\end{equation}
where $\Hessian[\cdot]$ is the Hessian matrix of the negative log-likelihood function. 
In our case, $\bfx$ is the camera pose and $\bfy$ is the rendered image at pose $\bfx$. $\bftheta$ represents the parameters of 3D Gaussians. Under the assumption of additive Gaussian noise, $p(\bfy \mid \bfx ; \bftheta) = \Normal\left(f(\bfx; \bftheta), \sigma^2I\right)$, where $f(\bfx; \bftheta)$ is the rendering function of 3DGS. In such case,
\begin{equation}
    \begin{aligned}
        \nabla_{\theta} \log p(\bfy \mid \bfx ; \bftheta) & = \frac{1}{\sigma^2} \nabla_{\theta} f(\bfx; \bftheta)^\top (\bfy - f(\bfx; \bftheta))
    \end{aligned}
\end{equation}
then the Fisher Information can be computed as:
\begin{equation}
    \begin{aligned}
        \Fisher(\bftheta)
         & = \Expect_{\bfY \sim p(\cdot \mid \bfx, \bftheta)}
        \left[
            \frac{1}{\sigma^4}
            \nabla_{\theta} f(\bfx; \bftheta)^\top
            (\bfY - f(\bfx; \bftheta))
            (\bfY - f(\bfx; \bftheta))^\top
            \nabla_{\theta} f(\bfx; \bftheta)
        \right]                                     \\
         & = \frac{1}{\sigma^2}\nabla_{\theta} f(\bfx; \bftheta)^\top
        \nabla_{\theta} f(\bfx; \bftheta)
    \end{aligned}
\end{equation}
The detailed derivation of Fisher Information is provided in Sec.~\ref{sec:fisher_derivation}.
Note that in our case, $\Fisher(\bftheta)$ is independent of $\bfy$, so Eq.~\ref{eq:fisher_hessian} can be simplified as:
\begin{equation}
    \Fisher(\bftheta)
    = \Expect_{\bfY \sim p(\cdot \mid \bfx; \bftheta)}
        \left[
            \Hessian[\bfY \mid \bfx; \bftheta]
        \right]
    = \Hessian[\bfy \mid \bfx; \bftheta]
\end{equation}

\section{Method}

\subsection{Overall Pipeline}
Fig.~\ref{fig:pipeline} illustrates the overall pipeline of our methods, which proceeds as follows: \textbf{(1) Initialization.} We initialize the 3DGS model and perform a brief warm-up optimization on the sparse-view training set. \textbf{(2) Stereo Augmentation with Fisher Information.} At the beginning of each epoch, we curate high-quality pseudo ground truths by warping the most informative training views with depth priors, and integrate them into the training set (Sec.~\ref{sec:fisher_augmentation}). \textbf{(3) Uncertainty-aware Regularization.} We optimize the 3DGS model on the augmented training set with our regularization, which adaptively adjusts the dropout probability of each Gaussian based on its optimization status (Sec.~\ref{sec:uncertainty_regularization}). Steps (2) and (3) are repeated iteratively until convergence.

\begin{figure}[h]
    \centering
    \includegraphics[width=1\linewidth]{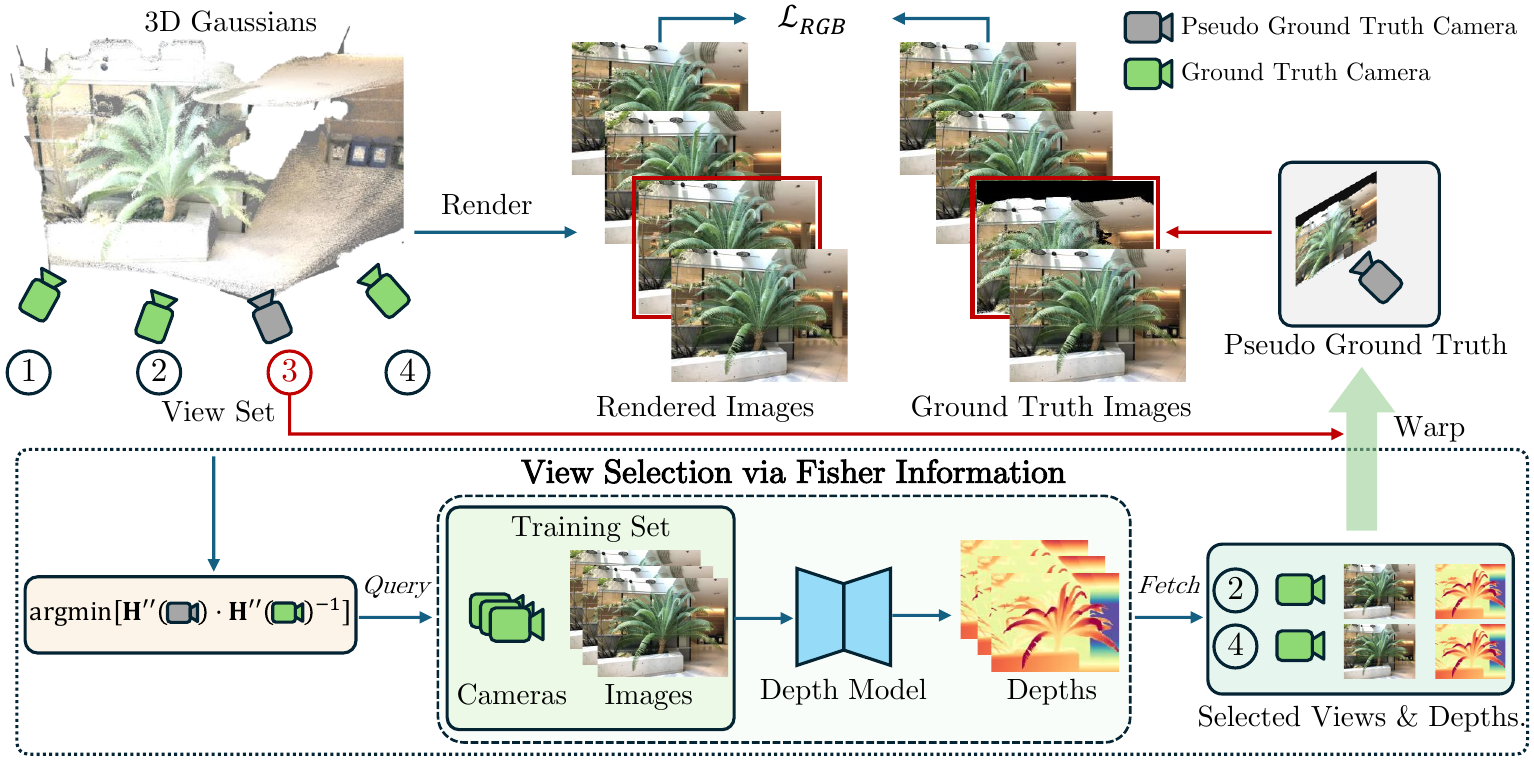}
    \vspace{-0.3cm}
    \caption{\textbf{Stereo Augmentation with Fisher Information.} With a sampled novel viewpoint, we select the most informative training views with the guidance of Fisher Information($\Hessian(\cdot)$). We then warp the selected ground truth views to the pseudo viewpoint with depth maps to curate high-quality pseudo ground truths.}
    \label{fig:augment}
    \vspace{-0.3cm}
\end{figure}

\subsection{Stereo Augmentation with Fisher Information}
\label{sec:fisher_augmentation}
We expands the training set by curating pseudo-ground truth views from existing observations and geometric priors. Specifically, we first sample a novel viewpoint. We then employ Fisher Information to select the most informative subset of training views relative to the novel viewpoint. Next, with depth maps produced by a pre-trained depth model \cite{depthanything3}, we warp the selected training views to the novel viewpoint to curate a high-quality pseudo ground truth. Finally, we integrate the pseudo ground truth into the training set to guide 3DGS optimization. The overall pipeline is illustrated in Fig~\ref{fig:augment}.

\mypar{Sampling Pseudo Ground Truth Viewpoints. }
We sample a novel viewpoint $\bfx'$ by randomly perturbing an existing training viewpoint $\bfx$. A viewpoint $\bfx$ is parameterized by an extrinsic matrix $E = [R \mid t] = [R \mid -R C]$, where $R \in \RealSet^{3 \times 3}$ is the rotation matrix, $t \in \RealSet^{3 \times 1}$ is the translation vector, and $C \in \RealSet^{3 \times 1}$ is the camera center in world coordinate system. We perturb the camera center $C$ by adding Gaussian noise $\epsilon_C \sim \Normal(0, \sigma_C^2 I)$. Then we perturb the camera orientation by smapling two random angles $\theta, \phi \sim \Normal(0, \sigma_\theta^2)$ to rotate the camera around the $x$- and $y$-axes respectively. We intentionally avoid rotation around the $z$-axis to prevent in-plane image distortion. The perturbed rotation matrix $R'$ is:
\begin{equation}
    R' =  R_x(\theta) R_y(\phi) R
\end{equation}
where $R_x(\theta), R_y(\phi)$ are axial rotation matrices. Finally, the extrinsic matrix $E'$ for pseudo ground truth viewpoint $\bfx'$ is computed as:
\begin{equation}
    \begin{aligned}
        E'
         & = [R' \mid t']                
         & = [R' \mid -R' (C + \epsilon_C)]
    \end{aligned}
\end{equation}


\mypar{Selecting Informative Support Views. }
To resolve the issue of \emph{Hollow and Random Augmentation}, we curate pseudo ground truths using multiple support views to make full use of geometric constraints. However, naively warping the whole set of training views degrades performance due to the defective depth predictions from distant views (detailed in Sec.~\ref{sec:ablation_stereo_augment}). Thus, we leverage Fisher Information to identify the most informative subset of training views with respect to the sampled viewpoint $\bfx'$.

Given the training set $\{(\bfx_k, \bfy_k)\}_{k=1}^K$ and the warmed-up rendering model $f(\bfx; \bftheta_0)$, we seek support views that maximize the similarity with the sampled novel view $(\bfx', \bfy')$. A naive measurement of similarity is Euclidean distance between camera centers; however, this ignores camera orientation and actual image captured, which leads to suboptimal view selection (see Sec.~\ref{sec:ablation_stereo_augment}). Instead, we adopt Information Gain (IG)~\cite{unifying_active_learning}, which inherently considers the actual visual similarity by measuring the reduction in model uncertainty. We define the IG between the sampled view $(\bfx', \bfy')$ and a candidate support view $(\bfx_k, \bfy_k)$ as:
\begin{equation}
    \IG[\Theta; \bfy' \mid \bfx', \bfx_k, \bfy_k]
    = \entropy[\Theta \mid \bfx_k, \bfy_k] - \entropy[\Theta \mid \bfx', \bfy', \bfx_k, \bfy_k]
\end{equation}
where $\entropy[\cdot]$ is the entropy function, and $\Theta \sim p(\bftheta)$ is the random variable of model parameters $\bftheta$.

Since directly computing the entropy is intractable, we employ Laplace approximation~\cite{unifying_active_learning} to simplify the computation of IG (detailed in Sec~\ref{sec:derivation}, Eq.~\ref{eq:ig_approx}) using the Fisher Information $\Hessian$, which gives:
\begin{equation}
    \begin{aligned}
        \IG[\Theta; \bfy' \mid \bfx', \bfx_k, \bfy_k]
         & = \frac{1}{2} \log \left| \Hessian[\bfy' \mid \bfx', \bftheta_0] \Hessian[\bftheta_0 \mid \bfx_k, \bfy_k]^{-1} + I \right| \\
         & \le \frac{1}{2} \trace(\Hessian[\bfy' \mid \bfx', \bftheta_0] \Hessian[\bftheta_0 \mid \bfx_k, \bfy_k]^{-1})
    \end{aligned}
    \label{eq:ig_trace}
\end{equation}
A lower IG indicates higher visual similarity between the views. Therefore, we identify the most informative support view by minimizing the trace upper bound of IG:
\begin{equation}
    k^* = \argmin_{k} \trace\left(\Hessian[\bfy' \mid \bfx', \bftheta_0] \Hessian[\bftheta_0 \mid \bfx_k, \bfy_k]^{-1}\right).
\end{equation}
To fully utilize multi-view geometric constraints rather than relying on a single view, we select all views whose IG falls within a threshold $\eta > 1$ relative to the minimum IG to form the support view set:
\begin{equation}
    \mathcal{S} = \left\{
    (\bfx_k, \bfy_k) \mid \IG[\Theta; \bfy' \mid \bfx', \bfx_k, \bfy_k] \le \eta \IG[\Theta; \bfy' \mid \bfx', \bfx_{k^*}, \bfy_{k^*}]
    \right\},
\end{equation}

\mypar{Curating Pseudo Ground Truths with Depth Priors. }
To obtain geometric priors, we feed the training set $\{(\bfx_k, \bfy_k)\}$ into a pre-trained stereo depth estimation model, Depth-Anything-3 \cite{depthanything3}, yielding depth maps $\{D_k\}$ for each training view. To curate pseudo ground truth, we unproject the support views $\mathcal{S}$ to a 3D point cloud using their corresponding depth maps. Then, we rasterize this point cloud from the pseudo training viewpoint $\bfx'$ to obtain the pseudo ground truth image $\bfy'$ using soft z-buffer rasterization (detailed in Sec.~\ref{sec:warping_soft_zbuffer}).

\mypar{Augmenting the Dataset.}
Once the pseudo ground truth $(\bfx', \bfy')$ is curated, we integrate it into the original training set:
\begin{equation}
    \data^{\text{aug}} = \data^{\text{train}} \cup \{(\bfx', \bfy')\}
\end{equation}
The 3DGS model is optimized on the augmented dataset $\data^{\text{aug}}$ subsequently. This multi-view geometric priors enforces spatial consistency and mitigates the \textit{Hollow and Random Augmentation} issue, improving the novel view rendering quality.

\subsection{Uncertainty-Aware Regularization}
\label{sec:uncertainty_regularization}
To address the \emph{Regularization without Selectivity} issue, we propose Uncertainty-aware Regularization strategy guided by Fisher Information. 
By quantitatively measuring the uncertainty of each Gaussian, we adaptively adjusting its dropout probability to selectively remove over-optimized Gaussians. We further introduce a \textit{soft-scale dropout} strategy that maintains the opacity of dropped Gaussians rather than completely zeroing them out.

\begin{wrapfigure}[12]{r}{0.57\textwidth}
    \vspace{-1.5\baselineskip} 
    \centering
    \includegraphics[width=\linewidth]{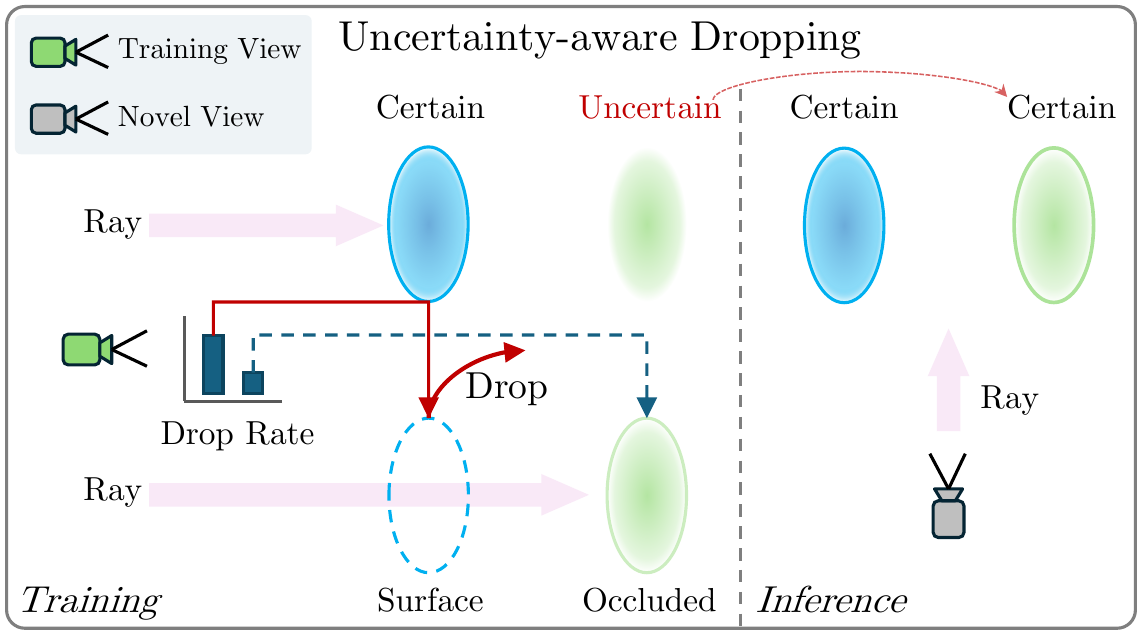}
    \vspace{-0.6cm}
    \caption{\textbf{Uncertainty-aware Dropout.}
    We selectively drop surface (over-optimized) Gaussians, while keeping occluded (under-optimized) ones.
    }
    \vspace{2\baselineskip}
    \label{fig:uncertainty_drop}
\end{wrapfigure}

\mypar{Uncertainty-aware Dropout.}
We quantify the optimization status of each 3D Gaussian $G_i$ using an uncertainty score $U_i$ derived from Fisher Information~\cite{pup3dgs} (illustrated in Fig.~\ref{fig:uncertainty_drop}).
Let $\theta_{G_i}$ denote the parameters of Gaussian $G_i$, we inspect the sub-matrix of Fisher Information $\Fisher_{G_i}$, which measures the sensitivity of the rendering function $f(\bfx; \bftheta)$ with respect to perturbations on Gaussian $G_i$:
\begin{equation}
    \Fisher_{G_i} = \nabla_{\theta_{G_i}} f(\bfx; \bftheta)^\top \nabla_{\theta_{G_i}} f(\bfx; \bftheta).
\end{equation}
Following PUP-3DGS~\cite{pup3dgs}, the uncertainty score $U_i$ is defined as:
\begin{equation}
    U_i = -\log \left|\nabla_{\theta_{G_i}} f(\bfx; \bftheta)^\top \nabla_{\theta_{G_i}} f(\bfx; \bftheta)\right|
\end{equation}
A higher $U_i$ indicates that $G_i$ is under-optimized, as perturbing it has minimal impact on the overall rendering model.

To prioritize the removal of over-optimized Gaussians, we propose a ranking-based selective dropout strategy. Given the target overall dropout probability $p$ and the number of Gaussians $N$, we firstly sort the Gaussians in ascending order of their uncertainty scores:
\begin{equation}
    \left\{ G_{(i)} \right\}_{i=1}^N
    \text{ such that }
    U_{(1)} \le U_{(2)} \le \ldots \le U_{(N)}
\end{equation}
we then construct a dropout candidate set $\mathcal{G}^{\text{cand}}$ containing $C \cdot p \cdot N$ Gaussians with the lowest uncertainty scores:
\begin{equation}
    \mathcal{G}^{\text{cand}} =
    \left\{
    G_{(i)} \mid i \le \lfloor C \cdot p \cdot N \rfloor
    \right\},
\end{equation}
where $1 < C < 1/p$ is a hyper-parameter to control the size of dropout candidate set. Finally, we apply dropout on the candidate set $\mathcal{G}^{\text{cand}}$ with dropout probability:
\begin{equation}
    p_{\text{cand}} = \frac{p}{C \cdot p} = \frac{1}{C}
\end{equation}
Such strategy ensures that over-optimized Gaussians face stricter regularization, while under-optimized Gaussians are preserved to undergo sufficient training.

\begin{wrapfigure}[11]{r}{0.66\textwidth}
    \vspace{-1.8\baselineskip} 
    \centering
    \includegraphics[width=\linewidth]{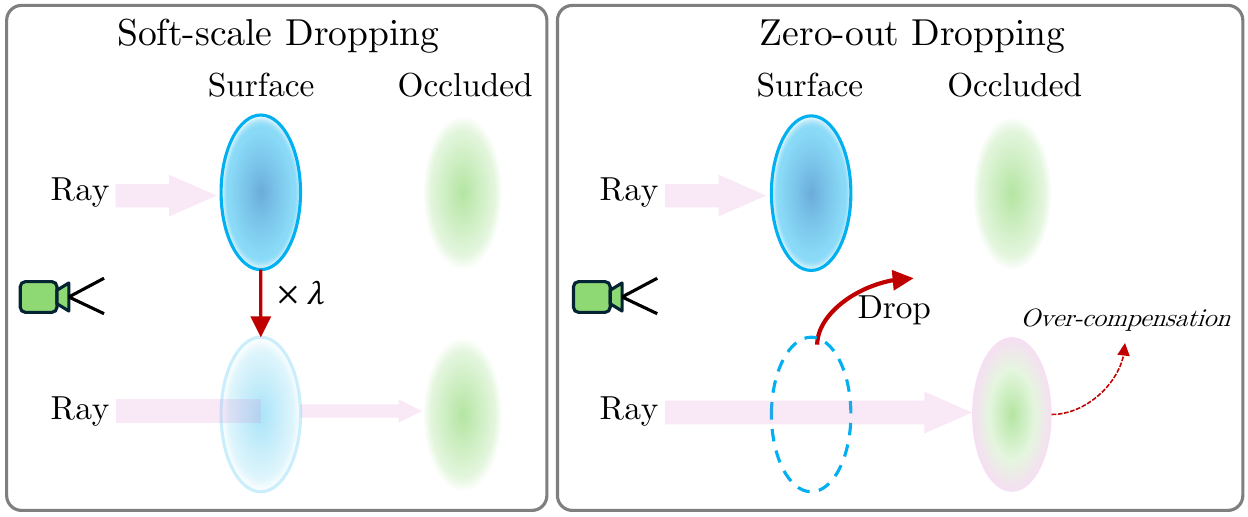}
    \vspace{-0.6cm}
    \caption{\textbf{Soft-scale Dropout.}
    We maintain the presence of Gaussian by scaling down their opacity with factor $\lambda$.
    }
    \label{fig:soft_dropout}
\end{wrapfigure}

\mypar{Soft-scale Dropout. }
Previous Dropout-based regularizations \cite{dropoutgs, dropgaussian} employ a hard removal strategy that zeros out the opacity of selected Gaussians. However, this removal triggers an \textit{over-compensation} effect (illustrated in Fig.~\ref{fig:soft_dropout}).
When a foreground Gaussian is completely removed, rays inadvertently penetrate to occluded region. Consequently, the optimization process forces the occluded Gaussians to \textit{over-compensation} by mimicking the missing foreground colors, which degrades the visual fidelity of novel view rendering.

To mitigate this artifact, we propose \textbf{Soft-scale Dropout}. Instead of completely removing the selected Gaussians $G_i$, we scale down its opacity $\alpha_i$ with a decay factor $\lambda \in (0,1)$:
\begin{equation}
    \alpha_i' = \lambda \alpha_i
\end{equation}
By maintaining non-zero opacity, the dropped Gaussians retain a soft presence in the rasterization and continue to contribute to the accumulated transmittance along the ray. This protects the occluded Gaussians from erroneous updates, and ensures that they are only fitted to the true background colors.

\input{tables/quantative_llff.tex}

\input{tables/quantative_dtu.tex}

\input{tables/quantative_mipnerf360.tex}

%% file: tables/quantative_llff.tex
\begin{table*}[t]
     \centering
     \small
     \caption{\textbf{Comparison with baseline methods on LLFF dataset.} Our method achieve the best performance across nearly all metrics in all settings. }
     \resizebox{\linewidth}{!}{
     \begin{tabular}{l|ccc|ccc|ccc}
          \toprule
          \multirow{2}{*}{Methods} & \multicolumn{3}{c|}{3 views} & \multicolumn{3}{c|}{6 views} & \multicolumn{3}{c}{9 views} \\
          \cmidrule(lr){2-4} \cmidrule(lr){5-7} \cmidrule(lr){8-10}
          & PSNR($\uparrow$) & SSIM($\uparrow$) & LPIPS($\downarrow$) & PSNR($\uparrow$) & SSIM($\uparrow$) & LPIPS($\downarrow$) & PSNR($\uparrow$) & SSIM($\uparrow$) & LPIPS($\downarrow$) \\
          \midrule
          RegNeRF\cite{regnerf} & 19.08 & 0.587 & 0.336 & 23.10 & 0.760 & 0.206 & 24.86 & 0.820 & 0.161 \\
          FreeNeRF\cite{freenerf} & 19.63 & 0.612 & 0.308 & 23.73 & 0.779 & 0.195 & 25.13 & 0.827 & 0.160 \\
          SparseNeRF\cite{sparsenerf} & 19.86 & 0.624 & 0.328 & - & - & - & - & - & - \\
          \midrule
          3DGS~\cite{3DGS} & 19.22 & 0.649 & 0.229 & 23.80 & 0.814 & 0.125 & 25.44 & 0.860 & 0.096 \\
          DNGaussian\cite{dngaussian} & 19.12 & 0.591 & 0.294 & 22.18 & 0.755 & 0.198 & 23.17 & 0.788 & 0.180 \\
          FSGS\cite{fsgs} & 20.43 & 0.682 & 0.248 & 24.09 & 0.823 & 0.145 & 25.31 & 0.860 & 0.122 \\
          CoR-GS\cite{cor_gs} & \cellcolor{tabthird}20.45 & \cellcolor{tabthird}0.712 & \cellcolor{tabsecond}0.196 & \cellcolor{tabthird}24.49 & \cellcolor{tabthird}0.837 & \cellcolor{tabfirst}0.115 & \cellcolor{tabthird}26.06 & \cellcolor{tabsecond}0.874 & \cellcolor{tabthird}0.089 \\
          DropGaussian\cite{dropgaussian} & \cellcolor{tabsecond}20.76 & \cellcolor{tabsecond}0.713 & \cellcolor{tabthird}0.200 & \cellcolor{tabsecond}24.74 & \cellcolor{tabsecond}0.837 & \cellcolor{tabthird}0.117 & \cellcolor{tabsecond}26.21 & \cellcolor{tabsecond}0.874 & \cellcolor{tabfirst}0.088 \\
          \textbf{Ours} & \cellcolor{tabfirst}21.12 & \cellcolor{tabfirst}0.729 & \cellcolor{tabfirst}0.190 & \cellcolor{tabfirst}24.91 & \cellcolor{tabfirst}0.840 & \cellcolor{tabsecond}0.116 & \cellcolor{tabfirst}26.27 & \cellcolor{tabfirst}0.875 & \cellcolor{tabfirst}0.088 \\
          \bottomrule
     \end{tabular}}
     \label{tab:quantative_llff}
\end{table*}

%% file: tables/quantative_dtu.tex
\begin{table*}[t]
     \centering
     \small
     \caption{\textbf{Comparison with baseline methods on DTU dataset.} Our method achieve the best performance across all metrics in all settings. }
     \resizebox{\linewidth}{!}{
     \begin{tabular}{l|ccc|ccc|ccc}
          \toprule
          \multirow{2}{*}{Methods} & \multicolumn{3}{c|}{3 views} & \multicolumn{3}{c|}{6 views} & \multicolumn{3}{c}{9 views} \\
          \cmidrule(lr){2-4} \cmidrule(lr){5-7} \cmidrule(lr){8-10}
          & PSNR($\uparrow$) & SSIM($\uparrow$) & LPIPS($\downarrow$) & PSNR($\uparrow$) & SSIM($\uparrow$) & LPIPS($\downarrow$) & PSNR($\uparrow$) & SSIM($\uparrow$) & LPIPS($\downarrow$) \\
          \midrule
          FreeNeRF \cite{freenerf} & 19.52 & 0.787 & 0.173 & 23.25 & 0.844 & 0.131 & \cellcolor{tabthird}25.38 & 0.888 & 0.102 \\
          SparseNeRF \cite{sparsenerf}  & 19.47 & 0.829 & 0.183 & -     &  -    &   -   & -     & -     & -     \\
          ReconFusion \cite{reconfusion} & 20.74 & 0.875 & 0.124 & 23.62 & 0.904 & 0.105 & 24.62 & 0.921 & 0.094 \\
          MuRF \cite{murf} & 21.31 & 0.885 & 0.127 & \cellcolor{tabthird}23.74 & \cellcolor{tabthird}0.921 & \cellcolor{tabthird}0.095 & 25.28 & \cellcolor{tabthird}0.936 & \cellcolor{tabthird}0.084 \\
          \midrule
          3DGS \cite{3DGS} & 10.99 & 0.585 & 0.313 & 20.33 & 0.776 & 0.223 & 22.90 & 0.816 & 0.173 \\
          FSGS \cite{fsgs} & 17.34 & 0.818 & 0.169 & 21.55 & 0.880 & 0.127 & 24.33 & 0.911 & 0.106 \\
          DNGaussian \cite{dngaussian} & 18.91 & 0.790 & 0.176 & 22.10 & 0.851 & 0.148 & 23.94 & 0.887 & 0.131 \\
          DropoutGS \cite{dropoutgs} & \cellcolor{tabthird}20.22 & \cellcolor{tabthird}0.830 & \cellcolor{tabthird}0.150 & - & - & - & - & - & - \\
          DropGaussian \cite{dropgaussian} & \cellcolor{tabsecond}21.44 & \cellcolor{tabsecond}0.888 & \cellcolor{tabsecond}0.103 & \cellcolor{tabsecond}25.58 & \cellcolor{tabsecond}0.926 & \cellcolor{tabsecond}0.073 & \cellcolor{tabsecond}27.65 & \cellcolor{tabsecond}0.950 & \cellcolor{tabfirst}0.052 \\
          \textbf{Ours} & \cellcolor{tabfirst}22.44 & \cellcolor{tabfirst}0.900 & \cellcolor{tabfirst}0.092 & \cellcolor{tabfirst}26.00 & \cellcolor{tabfirst}0.932 & \cellcolor{tabfirst}0.069 & \cellcolor{tabfirst}27.99 & \cellcolor{tabfirst}0.954 & \cellcolor{tabfirst}0.052 \\
          \bottomrule
     \end{tabular}}
     \label{tab:quantative_dtu}
\end{table*}

%% file: tables/quantative_mipnerf360.tex
\begin{table}[t]
     \centering
     \small
     \caption{\textbf{Comparison with baseline methods on Mip-NeRF360 dataset.} Our method achieve the best performance across nearly all metrics. }
     \resizebox{0.8\linewidth}{!}{
     \begin{tabular}{l|ccc|ccc}
          \toprule
          \multirow{2}{*}{Methods} & \multicolumn{3}{c|}{12 views} & \multicolumn{3}{c}{24 views}                                                                                   \\
          \cmidrule(lr){2-4} \cmidrule(lr){5-7}
          & PSNR($\uparrow$) & SSIM($\uparrow$) & LPIPS($\downarrow$) & PSNR($\uparrow$) & SSIM($\uparrow$) & LPIPS($\downarrow$) \\
          \midrule
          3DGS \cite{3DGS} & 18.52 & 0.523 & \cellcolor{tabthird}0.415 & 22.80 & 0.708 & 0.276 \\
          FSGS \cite{fsgs} & 18.80 & 0.531 & 0.418 & \cellcolor{tabthird}23.70 & \cellcolor{tabthird}0.745 & \cellcolor{tabthird}0.230 \\
          CoR-GS \cite{cor_gs} & \cellcolor{tabthird}19.52 & \cellcolor{tabthird}0.558 & 0.418 & 23.39 & 0.727 & 0.271 \\
          DropGaussian \cite{dropgaussian} & \cellcolor{tabsecond}19.74 & \cellcolor{tabsecond}0.577 & \cellcolor{tabsecond}0.364 & \cellcolor{tabsecond}24.13 & \cellcolor{tabsecond}0.762 & \cellcolor{tabfirst}0.225 \\
          \textbf{Ours} & \cellcolor{tabfirst}20.09 & \cellcolor{tabfirst}0.591 & \cellcolor{tabfirst}0.360 & \cellcolor{tabfirst}24.21 & \cellcolor{tabfirst}0.763 & \cellcolor{tabsecond}0.227 \\
          \bottomrule
     \end{tabular}}
     \label{tab:quantative_mipnerf360}
\end{table}

%% file: 4_experiment.tex
\section{Experiments}

\subsection{Dataset}

We evaluate our method on three widely adopted novel view synthesis benchmarks: LLFF \cite{llff}, Mip-NeRF 360 \cite{mip_nerf_360}, and DTU \cite{dtu}. Following previous methods, we train on 3-, 6-, and 9-views for both the LLFF and DTU datasets, and on 12- and 24-views for the Mip-NeRF 360 dataset. To be consistent with prior works, we downsample the original image resolutions by a factor of 8 for LLFF and Mip-NeRF 360, and by a factor of 4 for DTU.

\subsection{Metrics}

To quantitatively measure the rendering quality of our method, we adopt three widely adopted metrics, Peak Signal-to-Noise Ratio (PSNR), Structural Similarity Index Measure (SSIM), and Learned Perceptual Image Patch Similarity (LPIPS). PSNR accesses pixel-wise fidelity by average peak error between the rendered and ground truth images. SSIM evaluates the preservation of structural information, luminance, and contrast. LPIPS assesses perceptual quality by comparing deep feature representations.

\subsection{Quantitative Results}

\mypar{LLFF Dataset.}
As reported in Table~\ref{tab:quantative_llff}, our method establishes a new state-of-the-art on the LLFF dataset across 3-, 6-, and 9-view settings. In the challenging 3-view scenario, our method surpasses the previous best method by 0.36dB in PSNR, 0.016 in SSIM, and reduces LPIPS by 0.010. Our method also outperforms all NeRF-based and 3DGS-based methods in PSNR and SSIM under 6- and 9-view settings.

\mypar{Mip-NeRF 360 Dataset.}
We evaluate our method's performance in complex outdoor scenes on Mip-NeRF 360 dataset under 12- and 24-view settings. As shown in Table~\ref{tab:quantative_mipnerf360}, our method achieves the highest scores on nearly all metrics. Notably, in 12-view setting, our method outperforms the previous best method significantly (+0.35dB PSNR, +0.014 SSIM, -0.004 LPIPS), demonstrating the superiority of our method in handling complex outdoor scenes with sparse views.

\mypar{DTU Dataset.}
For object-centric reconstruction (Table~\ref{tab:quantative_dtu}), our method outperforms all baselines across all metrics under 3-, 6-, and 9-view settings. Specifically, in the 3-view scenario, our method achieves 22.44dB in PSNR, 0.900 in SSIM, and 0.092 in LPIPS, marking a significant leap over existing methods.

\input{tables/main_ablation.tex}

\subsection{Qualitative Results}

As shown in Fig~\ref{fig:llff_compare} and Fig~\ref{fig:mip_compare}, we present qualitative comparisons against state-of-the-art methods on LLFF and Mip-NeRF 360 dataset in 3-view and 12-view settings, respectively. Compared to Cor-GS and DropGaussian, our method renders novel views with overall better visual quality and less artifacts, which proves the effectiveness of our \textbf{Stereo Augmentation with Fisher Information} and  \textbf{Uncertainty-aware Regularization}.

\subsection{Ablation Study}

\mypar{Ablation of Proposed Components.}
Table~\ref{tab:main_ablation} quantitatively ablates the effectiveness of our proposed components on LLFF dataset with 3 views. \textbf{(1) Depth-based Augmentation (DBA).}
We evaluate it as a baseline augmentation strategy.
DBA curates pseudo ground truths by warping a single closest training view using depth priors produced by Depth-Anything-3 \cite{depthanything3}. This strategy effectively mitigates overfitting and improves the performance over the vanilla baseline (+0.41dB PSNR, +0.016 SSIM, and -0.006 LPIPS). \textbf{(2) Stereo Augmentation with Fisher Information (SAFI).} Based on DBA, we replace the single-view selection with our multi-view selection strategy guided by Fisher Information. This maximizes the utility of depth priors and suppresses errors from inaccurate depth estimations. Applying SAFI on top of DBA yields further improvements (+0.18dB PSNR, +0.002 SSIM, and -0.003 LPIPS). \textbf{(3) Uncertainty-aware Regularization (UAR).} Applying our uncertainty-aware dropout and soft-scale dropout to the baseline effectively regularizes the over-optimized Gaussians without causing visual artifacts. This standalone components improves PSNR by +0.16dB, SSIM by +0.009, and reduces LPIPS by 0.005. Integrating all proposed components, our method achieves the best performance (21.12dB PSNR, 0.729 SSIM, 0.190 LPIPS), proving the effectiveness of our method. Fig~\ref{fig:ablation} qualitatively demonstrates that each components progressively eliminates artifacts and enhances the fidelity of novel view synthesis. We provide more extensive ablation studies in Appendix (Sec.~\ref{sec:ablation_regularize}, Sec.~\ref{sec:ablation_stereo_augment}).

%% file: tables/main_ablation.tex
\begin{table}[t]
     \centering
     \small
     \caption{\textbf{Ablation study of our proposed components on LLFF dataset (3-view).} \textbf{DBA} means depth-based augmentation, \textbf{SAFI} means Stereo Augmentation with Fisher Information, and \textbf{UAR} means Uncertainty-aware Regularization.}
     \begin{tabular}
          {ccc|ccc}
          \toprule
          DBA & SAFI & UAR & PSNR($\uparrow$) & SSIM($\uparrow$)& LPIPS($\downarrow$) \\
          \midrule
          \multicolumn{3}{c|}{Baseline} & 20.46 & 0.706 & 0.202  \\
          \midrule
          \cmark &        &        & 20.87 & 0.722 & 0.196 \\
          \cmark & \cmark &        & 21.05 & 0.724 & 0.193 \\
          &  & \cmark & 20.62 & 0.715 & 0.197 \\
          \cmark & \cmark & \cmark & 21.12 & 0.729 & 0.190  \\
          \bottomrule
     \end{tabular}
     \vspace{-0.3cm}
     \label{tab:main_ablation}
\end{table}

%% file: 5_conclusion.tex
\section{Conclusion}

In this work, we propose a novel method for sparse-view 3DGS that leverages Fisher Information to guide the augmentation and regularization in training process. Our method uses Fisher Information to select the most informative views to curate high-quality pseudo ground truths with depth priors. It further leverages Fisher Information to measure the uncertainty of each 3D Gaussian and adaptively adjust the dropout probability for each Gaussian. Extensive experiments on three widely-used benchmarks demonstrate that our method achieves state-of-the-art performance in sparse-view 3DGS.